\documentclass[11pt]{article}

\usepackage[final]{acl}

\usepackage{times}
\usepackage{wrapfig}  
\usepackage{latexsym}
\usepackage{amsmath}
\usepackage{bm}
\usepackage{booktabs}
\usepackage[most]{tcolorbox}
\usepackage{hyperref}
\usepackage{url}
\usepackage{graphicx}
\usepackage{multirow} 
\usepackage{booktabs}
\usepackage{listings}
\usepackage{subcaption}
\usepackage{xcolor}
\usepackage{makecell}
\usepackage{enumitem}
\usepackage{tcolorbox}
\tcbuselibrary{breakable}
\tcbuselibrary{listings}
\usepackage{amssymb}

\usepackage{caption}
\usepackage{fancyvrb}
\usepackage{fvextra}

\definecolor{issueborder}{HTML}{15071A}
\definecolor{issuefill}{HTML}{F6F8FA}
\definecolor{envfill}{HTML}{FFF7F3}
\definecolor{envborder}{HTML}{D35B27}
\definecolor{agentfill}{HTML}{FBFCFE}
\definecolor{agentborder}{HTML}{0C69DA}
\definecolor{goldpatchborder}{HTML}{FABB00}
\definecolor{goldpatchfill}{HTML}{FFF7E1}
\definecolor{swecream}{RGB}{255,247,236}

\DefineVerbatimEnvironment{CodeVerbatim}{Verbatim}{
  formatcom={\color{black}},
  fontsize=\small,
  fontfamily=\ttdefault,
  fontseries=\mddefault,
  fontshape=\updefault,
  fillcolor=\color{white},
  framerule=0pt
}

\newtcolorbox{observationbox}[1][]{
        colback=envfill,
        colbacktitle=envfill,
        colframe=envborder,
        arc=5pt,
        fontupper=\small,
        fonttitle=\bfseries\color{black},
        boxrule=0.5mm,
        boxsep=1mm,
        width=\linewidth,
        breakable,
        title={User Output \hfill #1},
        rounded corners,
        toptitle=0.7mm,
        bottomtitle=0.7mm
}

\newtcolorbox{goldpatchbox}[1][]{
        colback=goldpatchfill,
        colbacktitle=goldpatchfill,
        colframe=goldpatchborder,
        arc=5pt,
        fontupper=\small,
        fonttitle=\bfseries\color{black},
        boxrule=0.5mm,
        boxsep=1mm,
        width=\linewidth,
        breakable,
        title={Output Patch \hfill #1},
        rounded corners,
        toptitle=0.7mm,
        bottomtitle=0.7mm
}

\newtcolorbox{issuebox}[1][]{
        colback=issuefill,
        colbacktitle=issuefill,
        colframe=issueborder,
        arc=5pt,
        fontupper=\small,
        fonttitle=\bfseries\color{black},
        boxrule=0.5mm,
        boxsep=1mm,
        width=\linewidth,
        breakable,
        title={Issue \hfill #1},
        rounded corners,
        toptitle=1mm
}

\newtcolorbox{agentbox}[1][]{
        colback=agentfill,
        colbacktitle=agentfill,
        colframe=agentborder,
        arc=5pt,
        fontupper=\small,
        fonttitle=\bfseries\color{black},
        boxrule=0.5mm,
        boxsep=1mm,
        width=\linewidth,
        breakable,
        title={mini-SWE-agent (Replayed Trajectory) \hfill #1},
        rounded corners,
        toptitle=1mm,
        lower separated=false
}

\newtcolorbox{agentboxInjected}[1][]{
        colback=red!10,
        colbacktitle=red!20,
        colframe=red,
        arc=5pt,
        fontupper=\small,
        fonttitle=\bfseries\color{black},
        boxrule=0.5mm,
        boxsep=1mm,
        width=\linewidth,
        breakable,
        title={mini-SWE-agent (Continued Generation) \hfill #1},
        rounded corners,
        toptitle=1mm,
        lower separated=false
}

\newtcolorbox{fileviewerbox}[1]{
        enhanced,
        breakable,
        boxrule = 1.5pt,
        fontupper = \small,
        fonttitle = \bf\color{black},
        arc = 5pt,
        rounded corners,
        colframe = black,
        colbacktitle = swecream,
        colback = swecream,
        title = #1,
        left=4pt, 
}

\usepackage{booktabs}   
\usepackage{multirow}   
\usepackage{makecell}   
\usepackage[table]{xcolor} 
\usepackage{graphicx}   
\usepackage{booktabs}
\usepackage{multirow}
\usepackage{makecell}
\usepackage[table]{xcolor}
\usepackage{graphicx}
\usepackage{tikz}
\usepackage{longtable}

\usetikzlibrary{positioning}
\usepackage[T1]{fontenc}
\usepackage[utf8]{inputenc}
\usepackage{microtype}
\usepackage{inconsolata}
\usepackage[table]{xcolor}
\usepackage{graphicx}
\usepackage{color-edits}
\usepackage{fancyhdr} 
\title{MedVerse: Efficient and Reliable Medical Reasoning via DAG-Structured Parallel Execution}

\author{
\textbf{Jianwen Chen\textsuperscript{1*},} \quad
\textbf{Xinyu Yang\textsuperscript{2*},} \quad
\textbf{Peng Xia\textsuperscript{1}} \quad
\textbf{Arian Azarang\textsuperscript{1}} \quad
\textbf{Yueh Z Lee\textsuperscript{1}} \quad \\
\textbf{Gang Li\textsuperscript{1}} \quad
\textbf{Hongtu Zhu\textsuperscript{1}} \quad
\textbf{Yun Li\textsuperscript{1}} \quad
\textbf{Beidi Chen\textsuperscript{2}} \quad
\textbf{Huaxiu Yao\textsuperscript{1}} \\
\textsuperscript{1}UNC-Chapel Hill \quad
\textsuperscript{2}Carnegie Mellon University \quad \\
\small{\textsuperscript{*}Equal contribution}
}

\begin{document}

\maketitle
\begin{abstract}
Large language models (LLMs) have demonstrated strong performance and rapid progress in a wide range of medical reasoning tasks.
However, their sequential autoregressive decoding forces inherently parallel clinical reasoning, such as differential diagnosis, into a single linear reasoning path, limiting both efficiency and reliability for complex medical problems.
To address this, we propose MedVerse, a reasoning framework for complex medical inference that reformulates medical reasoning as a parallelizable directed acyclic graph (DAG) process based on Petri Net theory.
The framework adopts a full-stack design across data, model architecture, and system execution.
For data creation, we introduce the MedVerse Curator, an automated pipeline that synthesizes knowledge-grounded medical reasoning path and transforms them into Petri Net–structured representations.
At the architectural level, we propose a topology-aware attention mechanism with adaptive position indices that supports parallel reasoning while preserving logical consistency.
Systematically, we develop a customized inference engine that supports parallel execution without additional overhead.
Empirical evaluations show that MedVerse improves strong general-purpose LLMs by up to 8.9\%. Compared to specialized medical LLMs, MedVerse achieves comparable performance with improved clinical reliability, while delivering a 1.3$\times$ reduction in inference latency and a 1.7$\times$ increase in generation throughput, enabled by its parallel decoding capability. Code is available at \href{https://github.com/aiming-lab/MedVerse}{https://github.com/aiming-lab/MedVerse}.

\end{abstract}

\section{Introduction}
\label{sec:introduction}

Recent advancements in Large Reasoning Models (LRMs)~\citep{openai2024learning,guo2025deepseek} have broadened the capabilities of medical artificial intelligence~\citep{moor2023foundation,thirunavukarasu2023large,nori2023capabilities,xia2024cares}, enabling a transition beyond simple information retrieval toward complex clinical reasoning~\cite{xia2024rule,xia2025mmed,alam2025towards,zhang2025patho}. 
In particular, state-of-the-art medical LRMs, such as MedReason~\citep{wu2025medreason}, HuatuoGPT-o1~\citep{chen2024huatuogpt}, and m1~\citep{huang2025m1}, have shown that Chain-of-Thought (CoT)~\citep{wei2022chain} reasoning can significantly enhance diagnostic accuracy. However, it remains unknown whether such models can be reliably and efficiently deployed in real-world clinical settings.

\begin{figure}[t]
    \centering
    \includegraphics[width=\linewidth]{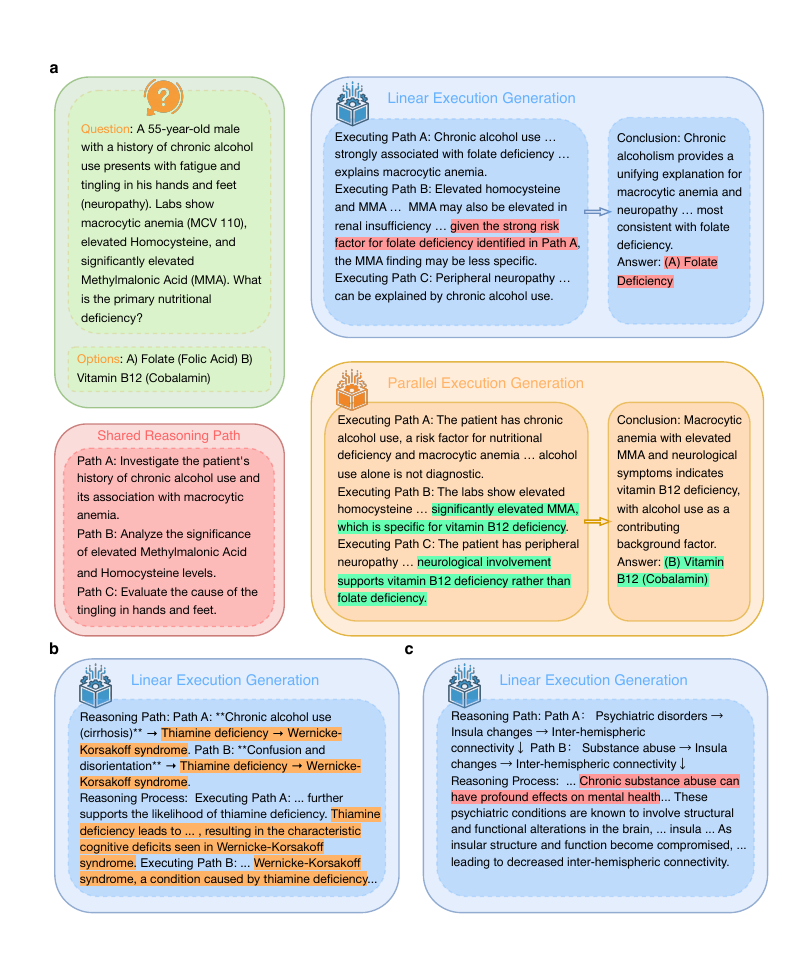}
    \caption{Limitations of sequential chain-of-thought reasoning in medical diagnosis. (a) \textbf{Accuracy:} Linear execution suffers from contextual pollution due to early incorrect hypotheses (\textcolor{red}{red}), whereas parallel reasoning preserves correct inference paths (\textcolor{green}{green}). (b) \textbf{Efficiency:} Sequential reasoning repeatedly processes overlapping evidence, leading to redundant computation (\textcolor{orange}{orange}). (c) \textbf{Interpretability:} Unstructured chain-of-thought (\textcolor{red}{red}) obscures explicit causal dependencies due to a structural mismatch with parallel clinical reasoning.
}
    \label{fig:4}
    \vspace{-1em}
\end{figure}

In practice, physicians increasingly employ these models to assist with clinical decision-making. Yet, the sequential nature of autoregressive (AR) models is misaligned with human cognitive processes, which naturally considers multiple differential diagnoses simultaneously~\citep{kassirer1978clinical}. Accordingly, this mismatch leads to three fundamental limitations: (i) \textit{Accuracy:} linear reasoning narrows the diagnostic hypothesis space, limiting the exploration of alternative diagnostic pathways; (ii) \textit{Efficiency:} serialized decoding increases response latency, posing challenges for real-time clinical decision support; (iii) \textit{Interpretability:} unstructured CoT lacks explicit causal relationships, hindering clinical interpretation and validation. An overview of these challenges is shown in Figure~\ref{fig:4}.

To address these bottlenecks, prior work on general-purpose LRMs has proposed parallel thinking~\citep{deepmind_gemini_imo2025, wang2025survey,ding2025dynamic}, enabling multiple reasoning branches to be explored concurrently before being synthesized into a final conclusion. However, most existing parallel generation frameworks induce parallelism primarily through brute-force repeat sampling at the early stages of generation~\citep{brown2024large}. Such approaches fail to capture the complex structure in clinical inference that involves multiple competing hypotheses and conditional dependencies~\citep{elstein1978medical}, resulting in suboptimal token efficiency and limited structural interpretability. Moreover, these inference-only solutions cannot internalize parallel thinking into the model, due to the absence of data-centric, end-to-end learning. While recent studies~\citep{yang2025multiverse, pan2025learning} enables recursive and consecutive parallel thinking patterns, they are limited to serial–parallel graph structures, failing to adapt either structure or knowledge required for complex clinical reasoning.

In this work, we overcome these challenges with \textbf{MedVerse}, a novel modeling framework that enables native directed acyclic graph (DAG)-based parallel generation tailored for medical reasoning. Specifically, we formalize the entire differential diagnosis process with Petri Net~\cite{petri1962kommunikation}, a bipartite graph composed of places and transitions. In this formulation, each place corresponds to a clinical entity (e.g., symptoms), while each transition encodes a directed relation between entities.

Built upon this structure, MedVerse is realized through a co-design of data, algorithm, and system:
\begin{itemize}[leftmargin=*]
\item \textbf{Data Curation.} We propose \textit{MedVerse Curator}, an automated  LLM-assisted pipeline that synthesizes knowledge-grounded medical reasoning trajectories. Starting from an entity-level knowledge graph extracted for each query, our curator leverages the Petri Net representation to transform it into a transition-level DAG, in which each node refers to an atomic reasoning step between entities. In practice, this process yields \textbf{MedVerse-14K}, a training corpus of 13,904 high-quality structured medical reasoning examples.
\vspace{-0.5em}
\item \textbf{Algorithm Design.} We propose \textit{MedVerse Attention}, a new attention mechanism for DAG-based execution. This is achieved by modifying attention masks and position embeddings to strictly ensure the DAG-structured dependency. This design also excels in data efficiency: since these changes are minor, pre-trained AR models
can be rapidly fine-tuned from causal attention to MedVerse attention through a few thousand examples.
\vspace{-0.5em}
\item \textbf{System Implementation.} We propose \textit{MedVerse Engine}, a high-performance serving engine built upon the Multiverse Engine~\cite{yang2025multiverse}. Our inference starts from a linear planning stage that generates the DAG-structured plan. Next, our engine dynamically extract this plan to enable true parallel decoding while respecting the underlying topological dependencies. Our engine enables parallel generation with negligible cost via continuous batching and radix attention.
\end{itemize}

Empirical evaluations on clinical benchmarks show that MedVerse improves accuracy by 4.8\% on Qwen2.5-7B and 8.9\% on Llama-3.1-8B, matching the performance of specialized reasoning models such as MedReason and HuatuoGPT-o1. Beyond accuracy, MedVerse overcomes the serial constraints of autoregression, achieving a 1.3$\times$ speedup in inference latency and a 69.3\% increase in peak throughput. 

\section{Related Work}
\label{sec:related_work}

\noindent \textbf{LLMs in Medical Reasoning.}
The application of LLMs in healthcare has evolved from general domain adaptation to specialized clinical reasoning. Early efforts, such as Med-PaLM~\citep{singhal2023large} and PMC-LLaMA~\citep{wu2023pmc}, focused on injecting medical knowledge via continual pre-training on biomedical corpora. Subsequently, instruction-tuned models like HuatuoGPT~\citep{zhang2023huatuogpt} and ChatDoctor~\citep{li2023chatdoctor} leveraged real-world dialogue data to align models with physician behaviors~\cite{li2023llava,moor2023med,nath2025vila}. More recently, the focus has shifted towards enhancing diagnostic logic, e.g., models trained on MedReason are explicitly supervised to generate CoT rationales~\cite{wu2025medreason,xia2025mmedagent,zhu2025mmedpo,lai2025med,huang2025medvlthinker}.
Despite these advances, existing models remain fundamentally constrained by the autoregressive decoding mechanism~\citep{vaswani2017attention}. They generate rationales as a rigid, serial token sequence, yielding an inference complexity of $\mathcal{O}(N)$. This computational linearity not only imposes high latency in real-time scenarios but also restricts the model to a single narrative thread, making it inefficient to maintain context across long, complex reasoning trajectories.

\noindent \textbf{Parallel Generation and Efficient Inference}
To break the serial decoding barrier, architectural approaches like Non-Autoregressive (NAR) decoding~\citep{gu2017non} and Speculative Decoding~\citep{leviathan2023fast} attempt simultaneous token generation, though often facing coherence or verification bottlenecks.
Notably, Multiverse introduces a MapReduce paradigm, decomposing generation into parallel ``Map'' branches merged via ``Reduce'' steps, supported by a specialized engine leveraging Radix Attention~\citep{zheng2024sglang} for efficiency.
However, these general-purpose paradigms lack medical grounding, and their topologies are constrained to simple fork-join patterns. This structural simplicity fails to capture the complex, non-linear branching networks of differential diagnosis~\citep{bowen2006educational}.
MedVerse addresses this limitation by co-designing a medically-grounded architecture with a custom vLLM-based engine~\citep{kwon2023efficient}, utilizing Radix Attention to enable zero-copy forking specifically tailored for complex medical reasoning topologies.
\section{MedVerse Modeling}
\label{sec:4}

This section introduces MedVerse, a novel modeling framework that reimagines the entire clinical reasoning process as DAG-based execution rather than sequential generation via a Petri Net abstraction, thereby improving reliability and efficiency.

\subsection{DAG Structure for Medical Reasoning}
\label{sec:dag_modeling}

As illustrated in Figure~\ref{fig:multiverse-petri}, multiple diagnostic hypotheses may share intermediate evidence or converge on common pathological mechanisms, leading to complex DAG-structured dependencies across different entities for medical reasoning. Such structures extend beyond the capabilities of sequential CoT reasoning or the tree-based extensions (e.g., Tree-of-Thoughts~\citep{yao2023tree}). Motivated by this observation, we formalize medical reasoning using a DAG $\mathcal{G} = (V, E)$ as follows:

\begin{itemize}[leftmargin=*]
    \item \textbf{Nodes as Reasoning States ($V$):} 
    Each node represents an intermediate reasoning state. Specifically, we distinguish three types of nodes: (i) \textit{source nodes}, corresponding to clinical entities grounded in the input question, which only have outgoing edges; (ii) \textit{hypothesis nodes}, representing diagnostic hypotheses or pathological states that may both \textit{split} into multiple downstream paths and \textit{merge} evidence from multiple upstream paths; and (3) \textit{conclusion nodes}, referring to final diagnostic outcomes, which only admit incoming edges and serve as unique convergence points of the entire reasoning process.
    
    \vspace{-0.5em}
    \item \textbf{Edges as Reasoning Steps ($E$):} 
    Each (directed) edges encode conditional dependencies between reasoning states. An edge $(u, v)$ indicates an admissible reasoning step from state $u$ to state $v$. To ensure structural acyclicity, all edges are restricted to forward dependencies that follow the temporal and causal order of clinical reasoning.
\end{itemize}

 Figure~\ref{fig:multiverse-petri} showcases that our formulation naturally captures DAG structures in clinical reasoning.

\subsection{Extension to Petri Nets}
\label{sec:preliminaries}

\begin{figure*}[t]
    \centering
    \includegraphics[width=\textwidth]{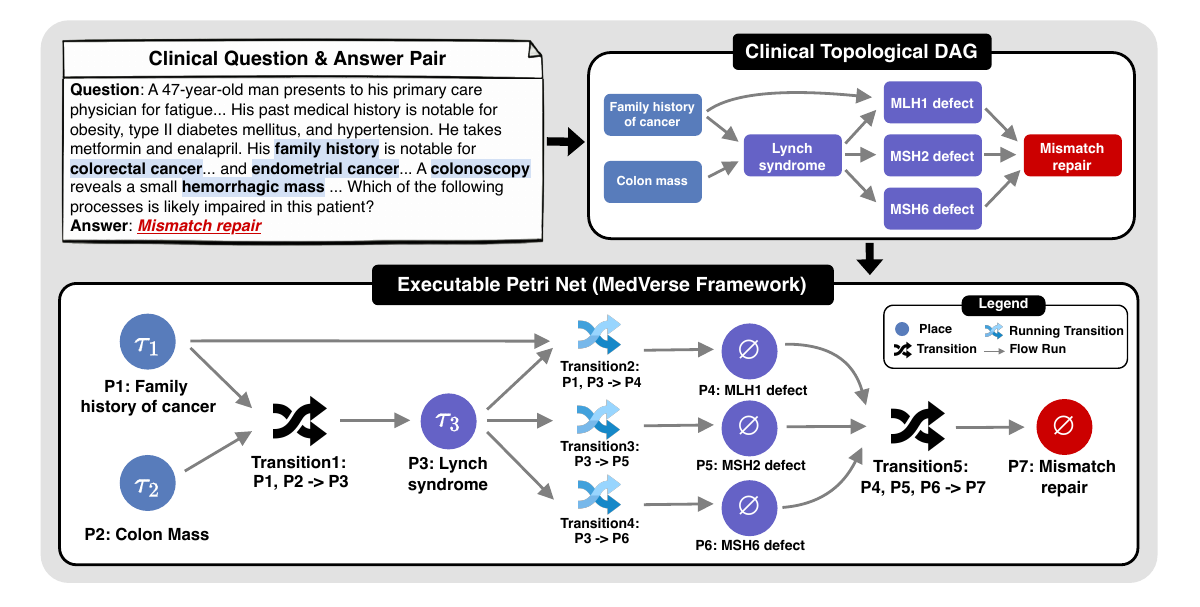}
    \caption{Illustration of the topological modeling process. The framework first extracts a structured clinical topological DAG from an unstructured question-answer pair, capturing causal dependencies. This DAG is then formally mapped into an Executable Petri Net, where reasoning states are instantiated as places and their dependency relations are realized through transitions.}
    \label{fig:multiverse-petri}
    \vspace{-1em}
\end{figure*}

While the DAG structure provides a topological blueprint, its static representation of solitary states fails to capture the generative transition logic essential for contextualizing history-dependent LLM inference. To realize this structure into an executable parallel process, we formalize our framework using Petri Nets. This mathematical grounding bridges the gap between the logical graph $\mathcal{G}$ and the physical execution of LLM inference, enabling reasoning states to be instantiated as places and causal dependencies as transitions for parallel execution (Figure~\ref{fig:multiverse-petri}).

\vspace{-0.5em}
\paragraph{Formal Definition.}
We define the execution model as a tuple $\mathcal{N} = (P, T, F, M_0)$, explicitly mapping to the DAG components:
\begin{itemize}[leftmargin=*, nosep]
    \item $P = \{p_1, \dots, p_m\}$ is a finite set of \textit{places}, corresponding to the nodes ($V$) in the DAG. Functionally, a place serves as a state container, holding the context or belief state waiting to be processed.
    \item $T = \{t_1, \dots, t_n\}$ is a finite set of \textit{transitions}, representing reasoning steps. Edges are mapped via many-to-one aggregation: converging edges (e.g., $A, B \to C$) form a single transition, while divergent edges ($A \to B, A \to C$) instantiate distinct transitions.
    \item $F \subseteq (P \times T) \cup (T \times P)$ defines the flow arcs that adhere to the DAG's topological direction. We denote the set of input places for a transition $t$ as the pre-set $\bullet t = \{p \mid (p,t) \in F\}$ and the output places as the post-set $t \bullet = \{q \mid (t,q) \in F\}$.
    \item $M_0$ is the initial marking such that for any place $p \in P$ corresponding to a DAG node with in-degree zero, $M_0(p)$ is non-empty, while $M_0(p)=\emptyset$ for all other places.
\end{itemize}

\vspace{-0.5em}
\paragraph{MedVerse Token Semantics.}
Standard Petri nets treat tokens as simple counters. To support the rich context of medical reasoning, we instantiate $\mathcal{N}$ as a Colored Petri Net (CPN). We define a token not as a scalar, but as a semantic tuple $\tau = (\mathbf{h}, \mathbf{k})$:
\begin{itemize}[leftmargin=*]
    \item $\mathbf{h}$: Encapsulates the textual history generated along the current path.
    \item $\mathbf{k}$: Denotes the KV-cache indices associated with that history.
\end{itemize}
This definition transforms tokens into computational state carriers, enabling the system to pass memory references rather than copying full text, which is pivotal for efficiency.

\paragraph{Execution as Inference.}
Inference is modeled as token flow through the Petri Net. A transition $t$ is enabled when tokens are present in all its input places and its output places are empty, ensuring each reasoning step executes exactly once. When fired, $t$ invokes the LLM to generate the corresponding reasoning output, reading from the input tokens and producing new tokens at its output places. Each resulting token inherits and extends both the textual history $\mathbf{h}$ and KV-cache references $\mathbf{k}$: the engine appends the newly generated text to $\mathbf{h}$ and maps the corresponding memory blocks to $\mathbf{k}$. Multiple enabled transitions may fire concurrently, yielding parallel decoding streams.

\subsection{Execution Semantics}
\label{sec:execution_paradigm}

Given the executable Petri Net constructed in Section~\ref{sec:preliminaries}, we specifies the execution semantics of MedVerse. We decouple \emph{scheduling} from \emph{execution}: scheduling identifies the set of causally enabled transitions ready for execution, while execution operates on this set via two abstract primitives, \textbf{Fork} and \textbf{Join}, to realize parallel generation with explicit causal synchronization.

\noindent \textbf{Scheduling via Enabled-Transition Frontier.}
At runtime, execution is governed by the current Petri Net marking $M_k$, which encodes token availability at each place. A transition is schedulable if tokens are present in all its input places and all output places are empty. The enabled-transition frontier at step $k$ is defined as
\begin{equation}\small
\begin{split}
    \mathcal{F}_k = \Big\{ t \in T \;\Big|\; & \big(\forall p \in \bullet t, M_k(p) \neq \emptyset\big) \; \land \\
    & \big(\forall q \in t\bullet, M_k(q) = \emptyset\big) \Big\},
\end{split}
\end{equation}
which represents the maximal set of transitions that can be executed concurrently without violating causal dependencies.

\noindent \textbf{Fork and Join Primitives.}
Given the frontier $\mathcal{F}_k$, the engine executes transitions based on their topological dependencies. \textbf{Fork} is applied to transitions sharing a common predecessor, initiating parallel decoding streams that inherit the exact same context. Conversely, \textbf{Join} is applied to transitions aggregating multiple reasoning branches, performing logical synchronization by merging distinct predecessor histories before generation proceeds.

\noindent \textbf{Execution Loop.}
After all transitions in $\mathcal{F}_k$ complete, the marking advances to $M_{k+1}$, the frontier is recomputed, and the scheduling–execution cycle repeats until no further transitions are enabled.

\subsection{Structured Generation Flow}
\label{sec:structured_flow}

\begin{figure}[h]
    \centering
    \includegraphics[width=\linewidth]{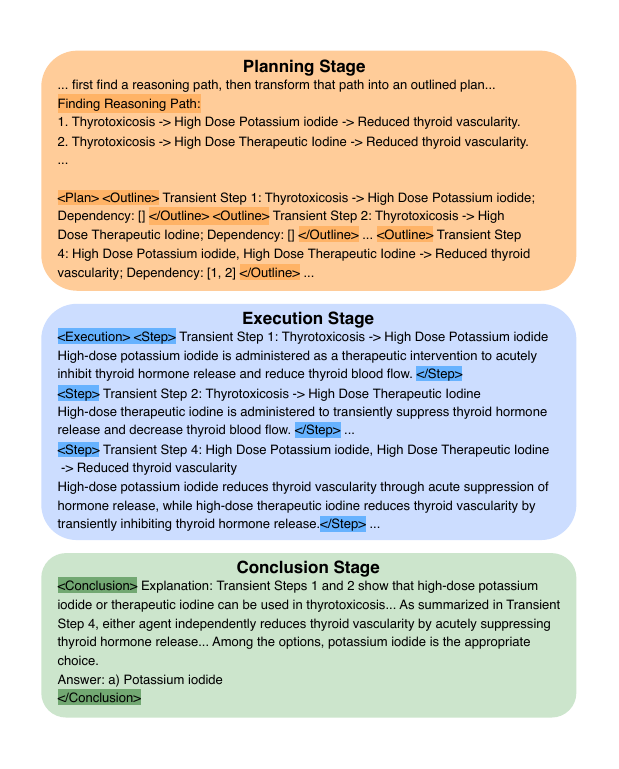}
    \caption{Example of the structured generation flow in MedVerse. The reasoning process is explicitly decomposed into planning, execution, and conclusion stages.}
    \label{fig:2}
    \vspace{-1em}
\end{figure}

Building on the structured representations and execution semantics introduced in Sections~3.1–3.3, MedVerse organizes LLM generation into a structured three-stage flow: planning, execution, and conclusion (see Figure~\ref{fig:2}). The process first constructs an explicit reasoning plan, executes it under graph-based constraints, and finally aggregates the results into a unified conclusion.

\noindent \textbf{Planning Stage.}
Generation begins with a planning stage that reconciles sequential generation with non-linear reasoning via a \emph{Think--then--Map} strategy. The model first generates multiple linear reasoning paths connecting clinical evidence to candidate diagnoses. These paths collectively encode overlapping dependencies that implicitly form a logical DAG. The model then consolidates them into a structured \texttt{<Plan>} block, where reasoning steps are organized using \texttt{<Outline>} tags annotated with dependency information. These annotations instantiate the flow relation $F$ of the Petri Net, specifying the reasoning topology while deferring detailed inference to execution.

\noindent \textbf{Execution Stage.}
Given the specified plan, generation proceeds within the \texttt{<Execution>} block following the execution paradigm in Section~\ref{sec:execution_paradigm}. At each iteration, the enabled-transition frontier is identified from the current marking. The engine then concurrently executes the transitions within this frontier by applying \textbf{Fork} and \textbf{Join}, encapsulating the generated reasoning for each transition within \texttt{<Step>} tags.

\noindent \textbf{Conclusion Stage.}
After all executable reasoning steps have completed, the process transitions to a final synthesis phase marked by the \texttt{<Conclusion>} block. In this phase, the outcomes of all completed reasoning branches are aggregated to produce a unified diagnostic conclusion.

\section{MedVerse Instantiation}
\label{sec:implementation}

To instantiate our MedVerse model discussed in Section~\ref{sec:4}, we present a comprehensive suite including three core components: the \textbf{MedVerse Curator} for structured data generation, the \textbf{MedVerse Attention} for DAG-structured modeling, and the \textbf{MedVerse Engine} to enable parallel execution.

\subsection{Data Curation: MedVerse Curator}
\label{sec:data_curation}

We introduce the MedVerse Curator, an LLM-assisted pipeline that extracts knowledge-grounded reasoning paths from a medical knowledge graph and compiles them into structured training instances executable under the MedVerse framework. Specifically, it includes four steps.

\noindent \textbf{I. Knowledge-Grounded Retrieval.}
The curator grounds medical reasoning in established clinical knowledge by first mapping questions and answer candidates to standardized medical concepts. It then retrieves plausible reasoning paths connecting these concepts through a medical knowledge graph, and prunes irrelevant or low-confidence branches.

\noindent \textbf{II. Topological Planning.}
The resulting linear reasoning skeletons are refined into executable plans using a specialized prompt that removes redundancy and enforces logical correctness and coherence. A DAG validity check ensures that all dependencies are acyclic; paths with invalid structures are discarded or re-routed.

\noindent \textbf{III. Structural Synthesis.}
Given a topology, the curator generates structured data in the MedVerse format. An LLM produces step-by-step reasoning for each execution transition, followed by an automated refinement module that smooths transitions across branches and ensures reasoning correctness. Finally, a coherent conclusion is synthesized from the refined execution trajectory.

\noindent \textbf{IV. Dual-Layer Verification.}
Finally, syntax-level validation and model-based logical evaluation are applied. Samples failing either check are iteratively regenerated until all constraints are satisfied.

\paragraph{MedVerse-14K Dataset.}
In practice, we apply our automated pipeline to the training subset of multiple medical training datasets, including MedQA~\citep{jin2021disease}, MedMCQA~\citep{pal2022medmcqa}, PubMedQA~\citep{jin2019pubmedqa}, MMLU-Medical~\citep{hendrycks2021ethics,hendryckstest2021}, HuatuoGPT-o1, MedXpert~\citep{zuo2025medxpertqa}, and Humanity’s Last Exam (HLE)~\citep{phan2025humanity}. This process yields \textbf{MedVerse-14K} that comprises 13,904 high-quality and structured reasoning trajectories, covering complex long-context medical problems.

\begin{table*}[!t]
\centering
\small
\setlength{\tabcolsep}{3.5pt} 
\caption{Performance comparison on medical reasoning benchmarks. We report accuracy (\%) across different datasets. \textbf{Bold} indicates the best performance.}
\label{tab:main_results}
\resizebox{\linewidth}{!}{\setlength{\tabcolsep}{1mm}{
\begin{tabular}{l | ccc | cccc}
\toprule
& \multicolumn{3}{c}{\textbf{Qwen2.5-7B-Instruct}} & \multicolumn{4}{|c}{\textbf{LLaMA-3.1-8B-Instruct}} \\
\cmidrule(lr){2-4} \cmidrule(lr){5-8}
\textbf{Benchmark} & Original & MedReason & \textbf{MedVerse} & Original & MedReason & Huatuo-o1 & \textbf{MedVerse} \\
\midrule
HLE (Biomed)     & 18.4 & \textbf{20.8} & 19.6 & 13.6 & 20.2 & 14.6 & \textbf{20.6} \\
MedBullets (op4)  & 45.8 & 49.7 & \textbf{55.2} & 48.7 & 57.1 & 55.8 & \textbf{62.3} \\
MedBullets (op5)  & 39.6 & 44.2 & \textbf{48.0} & 42.5 & 51.0 & \textbf{53.9} & 53.6 \\
MedQA      & 56.2 & 56.2 & \textbf{58.6} & 58.7 & 63.9 & \textbf{72.4} & 66.4 \\
MedXpert          & 12.3 & 14.5 & \textbf{15.3} & 13.2 & 18.4 & 16.8 & \textbf{19.3} \\
\midrule
\textbf{Average}  & 34.5 & 37.1 & \textbf{39.3} & 35.3 & 42.2 & 42.7 & \textbf{44.2} \\
\bottomrule
\end{tabular}}}
\vspace{-1.5em}
\end{table*}

\subsection{Algorithm Design: MedVerse Attention}
\label{sec:algorithm_design}

Next, we introduce \textbf{MedVerse Attention} to replace standard causal attention~\cite{vaswani2017attention}. The causal attention computes the $i$-th token's output with query \(\bm{q}_i\), and keys \(\bm{k}_j\), values \(\bm{v}_j\) (\(j \leq i\)):
\begin{equation}
\begin{split}
a_{ij}
&= \mathrm{Softmax}\Big(
(\bm{q}_i \cdot P(i))^\top (\bm{k}_j \cdot P(j)) + M_{ij}
\Big).
\end{split}
\end{equation}
where \(M_{ij} = 
\begin{cases}
0, & j \leq i \\
-\infty, & \text{otherwise}
\end{cases}\)
is casual attention mask and \(P(i)\) is embedding for  position $i$.

\noindent \textbf{DAG-based Attention Mask.} Transitions within the same enabled-transition frontier may execute concurrently and must remain causally independent. To enforce this property, we construct a layer-wise mutual exclusion mask $M$. During training, the input sequence is segmented into frontier layers according to the execution plan, where each layer contains a set of parallel reasoning steps ${S_1, S_2, \dots}$. For two tokens $i$ and $j$ associated with steps $S_u$ and $S_v$, the attention bias is defined as:
\begin{equation} \small
    M_{ij} =
    \begin{cases}
    -\infty & \text{if } j > i \\
    -\infty & \text{if } \mathrm{Layer}(i) = \mathrm{Layer}(j) \land S_u \neq S_v \\
    0 & \text{otherwise}
    \end{cases}
\end{equation}
The first condition preserves standard autoregressive causality, while the second prevents information leakage between parallel transitions.

\noindent \textbf{Adaptive Position Indices.} While masking enforces causal isolation, standard monotonic position indices cannot represent Petri Net execution synchronization. We therefore assign position indices based on the logical execution timeline rather than linear token order. Tokens generated by transitions within the same enabled-transition frontier are assigned an identical starting index (fork alignment). For transitions that join multiple branches in a subsequent frontier, the position index is set to the maximum index among all predecessor branches, allowing the model to attend to the complete causal history.

Together, topology-aware masking and adaptive position indices implement the execution semantics of Section~\ref{sec:execution_paradigm} within standard autoregressive Transformers, without modifying the training pipeline.

\subsection{Inference Engine: MedVerse Engine}
\label{sec:engine}

To translate the execution semantics of the Petri Net into practical inference speedups, we develop the MedVerse Engine, a high-performance serving system built upon the Multiverse Engine.
Unlike standard engines that treat generation as a uniform stream, our engine implements a Hybrid Execution Pipeline that seamlessly transitions from linear LLM-driven planning to system-guided parallel execution. This engine includes two phases, which are detailed as follows:

\paragraph{Phase I: Linear Planning \& Graph Initialization.}

The engine first employs standard autoregressive decoding to perform linear planning, during which the LLM translates the input into multiple linear reasoning paths followed by a topological \texttt{<Plan>} block. This phase is managed strictly as a conventional linear generation process to ensure logical correctness and coherence. Upon detecting the \texttt{</Plan>} tag, the engine pauses generation and parses the dependency annotations specified in the \texttt{<Outline>} tags to instantiate the in-memory Petri Net structure $\mathcal{N}$ and initialize the token marking $M_0$, thereby triggering an immediate transition to graph-based execution mode.

\paragraph{Phase II: Frontier-Based Graph Execution.}

Guided by the parsed topology, the scheduler identifies the enabled-transition frontier $\mathcal{F}_k$ (Sec.~\ref{sec:execution_paradigm}) at each step and executes all transitions in the frontier concurrently using two memory-optimized primitives. For transitions that branch from a common predecessor, the engine applies \emph{Fork} execution, spawning multiple parallel decoding streams that share the same prefix KV cache via Radix Attention, thereby enabling zero-copy prefix reuse until each corresponding \texttt{<Step>} block completes. Conversely, for transitions with multiple predecessors, execution is deferred until all upstream reasoning paths finish, after which the engine applies \emph{Join} execution by merging the KV states of all predecessor paths together with the preceding context to construct a unified KV cache. Leveraging the flexible radix cache layout, this merge is performed without padding or physical memory copying, and generation then proceeds from the merged KV state along the newly constructed sequence.

\section{Experiments}
\label{sec:5}

\subsection{Setup}
\label{sec:setup}
\textbf{Training.} We fine-tune MedVerse variants from the Qwen2.5-7B-Instruct~\citep{team2024qwen2} and Llama-3.1-8B-Instruct~\citep{dubey2024llama} checkpoints, incorporating our proposed MedVerse attention mechanism. Training is conducted on the curated MedVerse-14K dataset. All base models are trained for 3 epochs with a learning rate of $10^{-5}$ and a batch size of 128. Fine-tuning is conducted using 4 NVIDIA H200 GPUs with PyTorch FSDP.
\vspace{0.2em}

\noindent
\textbf{Evaluation.} We evaluate all MedVerse variants across five standard medical reasoning benchmarks, including MedQA, MedXpert, Humanity’s Last Exam (HLE)~\citep{phan2025humanity}, and MedBullets (op4, op5)~\citep{chen2025benchmarking}. All experiments are conducted using our SGLang-based MedVerse Engine.
\vspace{0.2em}

\noindent
\textbf{Baselines.} We compare MedVerse against the  original base models and other medical LLMs, including MedReason-8B and HuatuoGPT-o1-RL-8B. In addition to accuracy, we report both latency and throughput to assess efficiency gains of MedVerse. For fair comparisons, MedReason-8B is fine-tuned using the same number of examples as MedVerse-14K, while following its official training recipe.

\begin{figure}[h]
    \centering
    \includegraphics[width=\linewidth]{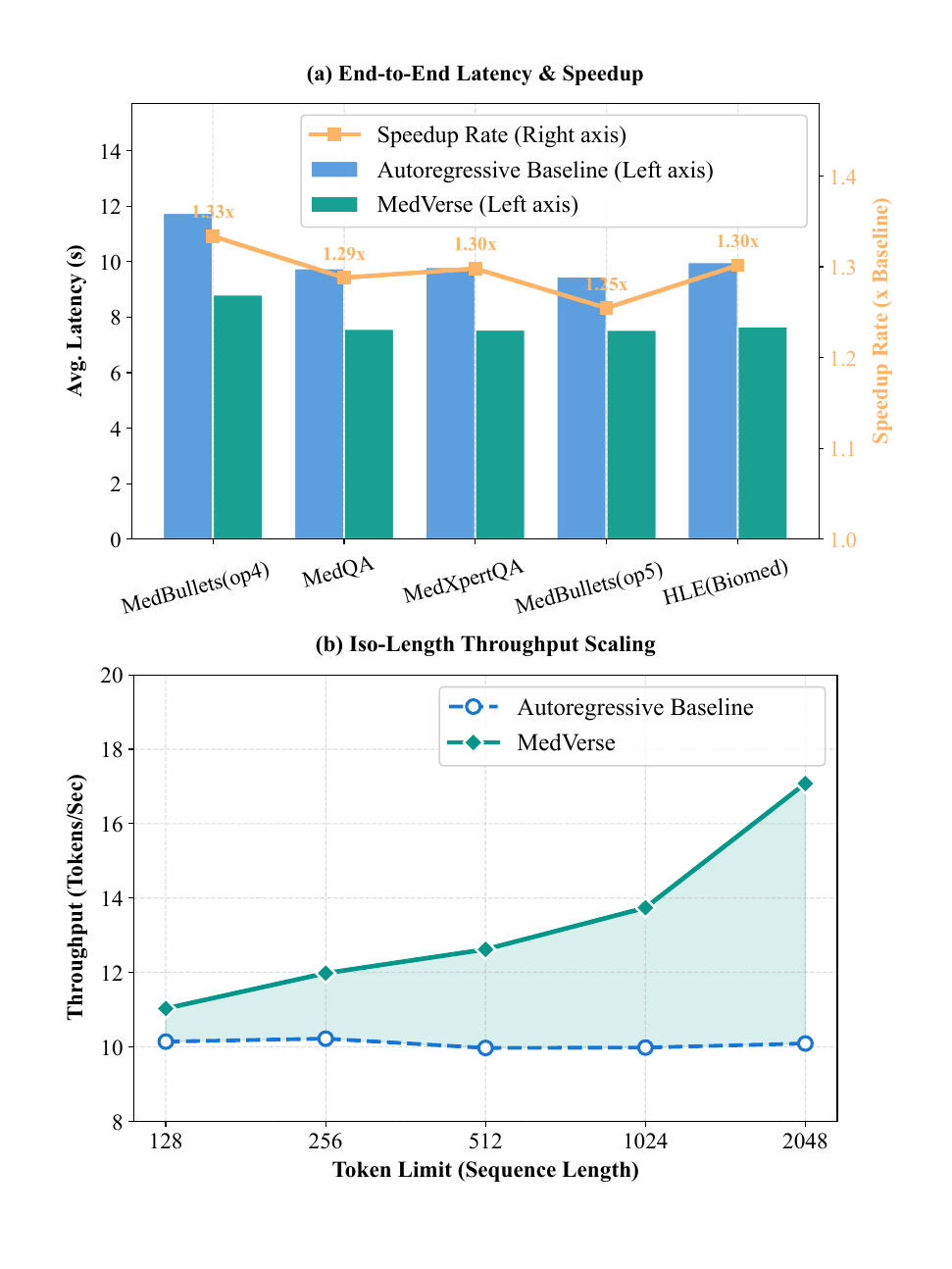}
    \caption{\textbf{Efficiency Metrics.} (a) Average latency and relative speedup (orange line) across five datasets. MedVerse consistently outperforms the baseline. (b) Throughput vs. Sequence Length. Our method exhibits better scaling properties, maintaining higher throughput as token complexity increases.}
    \label{fig:5}
    \vspace{-1.5em}
\end{figure}

\subsection{Performance Evaluation}
\label{sec:results_accuracy}

Table~\ref{tab:main_results} reports the performance of MedVerse on medical reasoning benchmarks. Across both backbones, MedVerse consistently outperforms the base models and strong autoregressive medical LLM baselines. For Qwen2.5-7B-Instruct, MedVerse improves the average accuracy from 34.5\% to 39.3\%, exceeding MedReason (37.1\%). For LLaMA-3.1-8B-Instruct, MedVerse achieves the highest average accuracy of 44.2\%, surpassing both MedReason (42.2\%) and HuatuoGPT-o1-RL-8B (42.7\%).
MedVerse further scales to 14B with a +18.0\% relative gain and consistent 1.32$\times$ latency speedup (Appendix~\ref{sec:scaling}).

\subsection{Efficiency Analysis}
\label{sec:efficiency}

To validate the efficiency of our system, we conducted a rigorous efficiency evaluation comparing MedVerse against standard autoregressive baselines on NVIDIA H200 GPUs. Our analysis focuses on three key dimensions: the latency for generating full reasoning chains, the engine's throughput under controlled output lengths, and the computational cost decomposition across pipeline stages.

\noindent \textbf{End-to-End CoT Latency.}
First, we measured the total wall-clock time required to answer medical queries across five diverse datasets (e.g., MedXpertQA, MedQA) with a batch size of 64. As shown in Figure \ref{fig:5}(a), MedVerse consistently outperforms the sequential AR baseline, achieving a stable speedup ranging from 1.25$\times$ to 1.33$\times$ (indicated by the orange trend line).
Theoretically, in standard AR models, generating a comprehensive diagnosis with multiple branches requires linear time $\mathcal{O}(N)$, where $N$ is the total token count. In contrast, MedVerse leverages its topological structure to generate independent reasoning paths simultaneously. This shifts the latency complexity from total length to the \textit{topological depth} of the reasoning graph $\mathcal{O}(D)$, drastically reducing user wait time for complex, multi-step queries.

\noindent \textbf{Iso-Length Throughput Comparison.}
To isolate the system's raw generation capability, we conducted an ``Iso-Length'' stress test on the HLE dataset with a batch size of 1, where both models were constrained to generate identical sequence lengths ranging from 128 to 2048 tokens.
Figure \ref{fig:5}(b) illustrates the throughput divergence. While the AR baseline maintains a relatively static throughput ($\sim$10 tokens/sec) limited by memory bandwidth and serial dependency, MedVerse demonstrates superior parallel scalability. The performance gap widens significantly as the sequence length increases: at 2048 tokens, our system achieves a peak throughput of $\sim$17.1 tokens/sec, representing a +69.3\% gain over the baseline. This confirms that MedVerse effectively converts the GPU's parallel compute capacity into valid token throughput, making it increasingly efficient for long-context medical reasoning tasks.

\begin{table}[ht]
\centering
\resizebox{\linewidth}{!}{
\begin{tabular}{lc}
\toprule
\textbf{Stage / Metric} & \textbf{Wall-clock Time (\%)} \\
\midrule
Planning Stage & 39\% \\
Execution Stage & 61\% \\
System Overhead (parsing \& scheduling) & {<}0.01\% \\
KV Fork/Join Cost (within Execution) & 1.1\% \\
\bottomrule
\end{tabular}}
\caption{\textbf{Computational Cost Decomposition.} Wall-clock time breakdown for MedVerse.}
\label{tab:cost}
\end{table}

\noindent \textbf{Computational Cost Decomposition.}
We further profile the wall-clock time distribution across pipeline stages (Table~\ref{tab:cost}) to isolate the sources of latency. The Planning Stage accounts for 39\% of total wall-clock time, while the Execution Stage dominates at 61\%. System-level parsing and scheduling overhead is negligible ($<$0.01\%), and the combined KV Fork/Join cost represents only 1.1\% of total wall-clock time, confirming that the observed speedups are not offset by infrastructure overhead. This is enabled by Radix Attention for $\mathcal{O}(1)$ Fork operations and zero-copy KV merges for Join primitives.

\subsection{Structural Analysis}
\label{sec:efficiency_breakdown}

To characterize when our approach benefits, we profile MedVerse with Qwen-2.5-7B across DAG topologies. In cases of a single linear chain, which comprise only 3\% of clinical cases, MedVerse maintains 1:1 latency parity with serial generation rather than degrading; the planning and parsing overhead is negligible at this scale. Significant efficiency gains are realized as DAG topologies becomes difficult (Table \ref{tab:topology}), shifting computational complexity from total token length to topological depth. This architectural alignment with non-linear medical reasoning ensures high reliability (up to 63.5\% accuracy) without computational friction.

\begin{table}[h]
    \centering
    \resizebox{\linewidth}{!}{
    \begin{tabular}{l ccc}
        \toprule
        \textbf{Topology} & \textbf{Prop.} & \textbf{Speedup} & \textbf{Acc. (\%)} \\
        \midrule
        Single Linear Chain   & 3\%  & 1.00$\times$ & -- \\
        Multi. Indep. Chains  & 58\% & \textbf{1.40$\times$} & 54.3 \\
        Complex Intersecting  & 39\% & \textbf{1.25$\times$} & \textbf{56.7} \\
        \bottomrule
    \end{tabular}}
    \caption{\textbf{Efficiency and Structural Analysis.} MedVerse yields meaningful speedups across 97\% of non-linear clinical cases while maintaining high accuracy.}
    \label{tab:topology}
\end{table}

\subsection{Clinical Reliability Analysis}
\label{sec:reliability}

To evaluate clinical reliability, we conducted a rigorous analysis using GPT-5.2 as a physician-level judge~\citep{zheng2023judging} across 50 cases from HLE (Biomed). We define five key metrics: Causal Validity (adherence to medical logic, 1--5 scale), Edge Accuracy (correctness of reasoning steps in the DAG, \%), Average Logical Jumps (unsupported transitions, average count per case), Mean Score (overall clinical utility, 1--5 scale), and High-Risk Error (frequency of dangerous guideline contradictions, categorized by levels 1/3/5). As shown in Table \ref{tab:reliability}, MedVerse demonstrates superior structural integrity; by anchoring reasoning branches in evidence via DAG-structured constraints, it significantly mitigates hallucinatory leaps. Notably, MedVerse reduces high-risk errors by 50\% compared to the serial baseline (5.7\% vs. 11.4\%), while improving causal validity by 12.1\% and edge accuracy by 15.4\%. These results confirm that our Join-based verification effectively enforces logical consistency in high-stakes clinical decision-making.

\begin{table}[h]
    \centering
    \small
    \begin{tabular}{lccc}
        \toprule
        \textbf{Metric} & \textbf{Serial} & \textbf{MedVerse} & \textbf{$\Delta$} \\
        \midrule
        Causal Validity $\uparrow$    & 1.82 & \textbf{2.04} & +12.1\% \\
        Edge Accuracy $\uparrow$     & 35.8 & \textbf{41.3} & +15.4\% \\
        Logical Jumps $\downarrow$    & 3.30 & \textbf{2.46} & -25.5\% \\
        Mean Score $\uparrow$         & 2.00 & \textbf{2.36} & +18.0\% \\
        High-Risk Error $\downarrow$  & 11.4 & \textbf{5.7}  & -50.0\% \\
        \bottomrule
    \end{tabular}
    \caption{\textbf{Clinical Reliability Evaluation} on 50 HLE (Biomed) cases. $\uparrow$ ($\downarrow$) indicates higher (lower) is better.}
    \vspace{-1.5em}
    \label{tab:reliability}
\end{table}

\subsection{Ablation Study}
\label{sec:ablation}

We conduct a targeted ablation to assess the necessity of linear-to-parallel hybrid reasoning by comparing MedVerse with a \textbf{Direct Petri Net} variant that directly generates topological structures without linear planning. As shown in Table~\ref{tab:ablation_topology}, when evaluated on MedXpert (batch size 1), the Direct Petri Net variant exhibits a substantial accuracy drop and underperforms even the autoregressive baseline, indicating that standard LLMs struggle to construct sound execution graphs from scratch. In contrast, MedVerse achieves both the highest accuracy (19.3\%) and the lowest latency (4.0s), demonstrating that linear planning is essential for reliable graph construction and that its combination with parallel execution yields the best accuracy–efficiency trade-off.

\begin{table}[t]
    \centering
    \resizebox{\linewidth}{!}{
        \begin{tabular}{lcccc}
            \toprule
            \textbf{Model Variant} & \textbf{Linear} & \textbf{Parallel} & \textbf{Accuracy (\%)} & \textbf{Latency (s)} \\
            \midrule
            Autoregressive & \checkmark & $\times$ & 18.4 & 5.1 \\
            Direct Petri Net        & $\times$   & \checkmark & 17.4 & 4.5 \\
            \textbf{MedVerse} & \checkmark & \checkmark & \textbf{19.3} & 4.0 \\
            \bottomrule
        \end{tabular}
    }
    \caption{\textbf{Efficacy of Linear-to-Parallel Hybridization.} Ablation study on the MedXpert benchmark.}
    \vspace{-1.5em}
    \label{tab:ablation_topology}
\end{table}

\noindent We further disentangle architectural from data contributions in Appendix \ref{sec:training_strategy}: under identical MedVerse-14K data, the DAG-based attention and parallel inference yield a +2.4\% gain over the standard autoregressive baseline (Table \ref{tab:ablation_modes}), confirming that the improvements are attributable to the architectural design rather than data exposure alone.
\section{Conclusion}
\label{sec:conclusion}

This work addresses the fundamental mismatch between sequential autoregressive decoding and the inherently parallel nature of clinical reasoning. By reformulating medical inference as a DAG-structured process grounded in Petri Net theory, MedVerse enables large language models to reason over multiple diagnostic hypotheses concurrently while preserving causal consistency. This paradigm alleviates key limitations of linear chain-of-thought reasoning in accuracy, efficiency, reliability, and interpretability, and offers a principled path toward structurally aligned medical reasoning systems suitable for real-world clinical decision support.

\section{Ethical Considerations}
This work focuses on methodological advances in structured and parallel reasoning for medical language models and does not involve new data collection from human subjects. All training and evaluation data are derived from existing, publicly available benchmarks or automatically generated synthetic reasoning trajectories. No personally identifiable information is used. While the proposed framework aims to improve reasoning accuracy, efficiency, and interpretability, it is not intended for direct clinical deployment without human oversight. Any real-world medical application should ensure appropriate clinical validation and adhere to established ethical and regulatory standards.

\section*{Limitations}
\label{sec:limitations}

While MedVerse demonstrates consistent gains in both reasoning accuracy and inference efficiency, several limitations merit discussion.
First, although MedVerse reduces inference latency by shifting computational complexity from sequence length to topological depth, the achievable speedup depends on the inherent parallelism of the reasoning graph. For reasoning problems with predominantly linear dependency structures, the benefits of parallel execution may be limited. Automatically adapting execution strategies to different reasoning topologies remains an important direction for future work.
Second, our evaluation focuses on offline benchmarks under controlled inference settings. While the proposed framework is designed with real-time clinical deployment in mind, additional studies are required to assess its behavior under interactive, user-in-the-loop scenarios, where user feedback, partial responses, or dynamic query refinement may influence the execution process.

\section*{Acknowledgement}
This work is partially supported by NSF \#2449442 and NIH R01AG085581, R01AG079291, R01AR083790 and P50HD103573. The Authors acknowledge the National Artificial Intelligence Research Resource (NAIRR) Pilot, NCSA DeltaAI and OpenAI API for contributing to this research result.

\bibliography{custom}

\appendix

\appendix
\clearpage


\section{Additional Analysis}
\subsection{Effect of Training Data Scale}

\begin{table}[h]
    \centering
    \resizebox{\linewidth}{!}{
        \begin{tabular}{lccccc}
            \toprule
            \textbf{Samples}           & \textbf{1k} & \textbf{2k} & \textbf{5k} & \textbf{8k} & \textbf{14k} \\
            \midrule
            \textbf{Subset Ratio}      & $\sim$7\% & $\sim$14\% & $\sim$36\% & $\sim$57\% & 100\% \\
            \textbf{Average Accuracy}  & 29.2\%      & 32.5\%      & 37.5\%      & 38.7\%      & \textbf{39.2\%} \\
            \bottomrule
        \end{tabular}
    }
    \caption{\textbf{Scalability Analysis.} The monotonic improvement in average accuracy across all benchmarks confirms the effectiveness of our data scaling strategy.}
    \label{tab:ablation_scale}
\end{table}

We investigate the scalability of our approach by fine-tuning the base model on varying subsets of the MedVerse-14K dataset (ranging from 1k to 14k samples).
As presented in Table \ref{tab:ablation_scale}, we observe a clear monotonic positive correlation between dataset size and the average reasoning accuracy across all evaluation benchmarks.
Notably, the model exhibits exceptional data efficiency: with only 5k samples ($\sim$36\% of the total data), it achieves an accuracy of 37.5\%, recovering over 95\% of the peak performance.
Furthermore, the performance does not plateau even at the 14k mark. This validates the high quality of our synthesized data and suggests that further scaling could yield even greater improvements.

\subsection{Scalability Across Model Sizes}
\label{sec:scaling}

To assess the scalability of MedVerse beyond 7B and 8B scales, we evaluate MedVerse fine-tuned on the Qwen2.5-14B-Instruct backbone. As shown in Table~\ref{tab:scaling}, \textbf{MedVerse-14B} achieves 44.74\% average accuracy, representing a \textbf{+18.0\%} relative gain over the official 14B baseline---exceeding the relative gain observed at the 7B scale (+14.2\%). This demonstrates that our DAG-structured inductive bias becomes increasingly effective as parameters increase. Crucially, MedVerse maintains a consistent 1.32$\times$ inference latency speedup at the 14B scale. As it is implemented via topology-aware attention masking and adaptive position indices, the framework remains compatible with larger Transformers and MoE architectures.

\vspace{0.3em}
\begin{table}[ht]
    \centering
    \resizebox{\linewidth}{!}{
    \begin{tabular}{lcccccc}
        \toprule
        \textbf{Model} & \textbf{HLE} & \textbf{MB-op4} & \textbf{MB-op5} & \textbf{MedQA} & \textbf{MedXpert} & \textbf{Avg.} \\
        \midrule
        Qwen2.5-7B (Base) & 18.4 & 45.8 & 39.6 & 56.2 & 12.3 & 34.46 \\
        MedVerse-7B & 19.6 & 55.2 & 48.0 & 58.6 & 15.3 & 39.34 \\
        \midrule
        Qwen2.5-14B (Base) & 10.7 & 54.2 & 49.0 & 61.7 & 13.9 & 37.91 \\
        MedVerse-14B & 26.2 & 62.3 & 52.0 & 64.0 & 19.2 & \textbf{44.74} \\
        \bottomrule
    \end{tabular}}
    \caption{\textbf{Scaling Performance Analysis.} MedVerse-14B achieves +18.0\% relative gain, exceeding the 7B-scale gain (+14.2\%), confirming that DAG-structured reasoning scales effectively.}
    \label{tab:scaling}
\end{table}

\subsection{Analysis of Training Strategies and Inference Modes}
\label{sec:training_strategy}

To further investigate the source of MedVerse's performance gains, we conducted a comprehensive ablation study decoupling the training strategies (standard autoregressive vs. MedVerse attention) from the inference execution mode (serial vs. parallel). We evaluated four configurations on the Qwen2.5-7B-Instruct backbone across our benchmarks. The aggregated results are summarized in Table~\ref{tab:ablation_modes}.

\noindent The four configurations are defined as follows.

\noindent \textbf{Auto-Ser}: Standard autoregressive training executed with a standard serial engine (baseline).

\noindent \textbf{Auto-Par}: Standard autoregressive training executed with our DAG-based parallel engine.

\noindent \textbf{Mask-Ser}: MedVerse topology-aware training (masked attention) executed serially.

\noindent \textbf{Mask-Par}: MedVerse topology-aware training executed with our DAG-based parallel engine.

\begin{table}[h]
    \centering
    \resizebox{\linewidth}{!}{
    \begin{tabular}{lcccc}
        \toprule
        \textbf{Metric} & \textbf{Auto-Ser} & \textbf{Auto-Par} & \textbf{Mask-Ser} & \textbf{Mask-Par (Ours)} \\
        \midrule
        Average Accuracy (\%) & 36.9 & 37.9 & 38.6 & \textbf{39.3} \\
        \bottomrule
    \end{tabular}
    }
    \caption{Ablation study on Training Strategies and Inference Modes. We report the average accuracy across five medical reasoning benchmarks (HLE, MedBullets-op4/op5, MedQA, MedXpert).}
    \label{tab:ablation_modes}
\end{table}

\paragraph{Discussion.}
The results demonstrate a strong synergistic effect, with Mask-Par achieving the highest accuracy (39.34\%, +2.44\% over baseline). This improvement highlights two critical insights. First, the superiority of Mask-Ser over the baseline (+1.66\%) confirms that topology-aware masking enhances learning; by ``hiding'' irrelevant parallel branches, the mechanism prevents reliance on spurious positional correlations and forces the model to focus on genuine causal dependencies. Second, regarding topological alignment, the consistent gains from parallel structures in both training and inference indicate that medical reasoning inherently follows a DAG topology rather than a linear chain. MedVerse succeeds precisely by aligning the computational framework with this non-linear cognitive structure.

\section{Detailed MedVerse Curator Pipeline}
\label{app:curator}

This appendix provides a detailed description of the MedVerse Curator, including phase-wise procedures for knowledge grounding, topological planning, structural synthesis, and data verification.

\paragraph{Phase 1: Knowledge-Grounded Path Initialization.}
The pipeline begins by anchoring the reasoning in established medical knowledge, leveraging the retrieval methodology from MedReason (Steps 1--3):
\begin{enumerate}[label=(\roman*), nosep]
    \item \textbf{Knowledge Retrieval:} We first retrieve potential reasoning paths connecting the question entities to answer candidates from a large-scale medical Knowledge Graph (KG).
    \item \textbf{Entity Mapping:} We perform medical entity extraction to map unstructured query terms to standardized KG nodes.
    \item \textbf{Path Pruning:} We search for and prune irrelevant branches to isolate the original, raw reasoning paths.
\end{enumerate}
This phase provides a "ground truth" skeleton, ensuring the subsequent generation is factually grounded rather than hallucinated.

\paragraph{Phase 2: Topological Planning and Filtering.}
We then transform these linear skeletons into the MedVerse architecture (Steps 4--5). We employ a specialized prompt to \textbf{filter and edit} the raw paths, removing redundancy and ensuring logical coherence. Crucially, we perform a \textbf{DAG Validity Check}: the refined path is analyzed to ensure it forms a valid Directed Acyclic Graph. If the dependencies form a cycle
, the path is rejected or re-routed.

\paragraph{Phase 3: Structural Synthesis and Refinement.}
This core phase generates the XML-structured data (Steps 6--8):
\begin{itemize}[leftmargin=*, nosep]
    \item \textbf{Plan Generation (\texttt{<Plan>}):} We iteratively decompose the verified path, utilizing regex-based formatting to outline the Petri Net structure, defining transitions and explicit dependency lists.
    \item \textbf{Parallel Execution (\texttt{<Execution>}):} The teacher LLM generates the "transient step" content for each transition.
    \item \textbf{Iterative Refinement:} We implement a \textbf{Refinement Module} to knit these independent steps into a cohesive narrative. This module deduplicates overlapping logic across parallel branches, removes non-contributory details, and smooths the transition flow to ensure the reasoning naturally bridges the gap from the problem description to the final entity.
    \item \textbf{Conclusion Synthesis:} Finally, conditioned on the refined execution trajectory, the model generates the \texttt{<Conclusion>}, providing the final answer and a holistic explanation.
\end{itemize}

\paragraph{Phase 4: Dual-Layer Verification Loop.}
To guarantee data quality, we append two supplementary validation stages:
\begin{enumerate}[label=(\alph*), nosep]
    \item \textbf{Syntax Verification:} Ensures strict adherence to the XML schema and Petri Net definition (e.g., matching \texttt{<Step>} indices to \texttt{<Outline>} plans).
    \item \textbf{Logic \& Completeness Evaluation:} An evaluator model assesses the reasoning chain for logical gaps and verifies that the conclusion correctly addresses the user goal.
\end{enumerate}
Data failing either check triggers an iterative regeneration loop until all criteria are met.

\section{Prompting Protocol for the MedVerse Curator}

In this section, we present the complete five-stage prompting protocol used by the MedVerse Curator to construct the MedVerse-14K dataset, powered by the GPT-5.1 model  accessed via the ChatGPT API. The goal of this protocol is to systematically transform question–answer pairs into structured, parallel medical reasoning trajectories suitable for topology-aware execution.

The protocol is implemented as an offline data construction pipeline consisting of five sequential phases. Phase~1 adopts the knowledge-grounded retrieval procedure proposed in MedReason to initialize medically valid reasoning paths; as this phase directly reuses an existing method without modification, we do not reproduce the original prompts. Phases~2--4 introduce new structured transformations that progressively induce parallel reasoning structure, repopulate detailed clinical content, and enforce execution-ready constraints. The full specifications and prompts for Phases~2--4 are listed below.

\begin{table*}[t]
\centering
\begin{tcolorbox}[
  title={Phase 2: Reasoning Chain Filtering Template},
  colback=gray!5,
  colframe=black!75,
  fonttitle=\bfseries,
  width=\textwidth,
  boxrule=0.4pt
]
\small

\textbf{\textsc{System Prompt}}

You are a strict reasoning chain filter. Given a question, a list of candidate reasoning chains (\texttt{original\_reasoning}), and the correct answer, select only the reasoning chains that are directly relevant for deriving the answer from the question.

\textbf{Filtering Rules (Follow All Exactly):}
\begin{enumerate}[label=\arabic*), leftmargin=*, nosep]
    \item \textbf{Relevance:} Keep only chains that directly or critically contribute to deriving the answer from the question. Discard any chain that is unrelated or unnecessary for reaching the answer.
    \item \textbf{Consistency:} Remove chains that contradict the facts stated in the question or that lead to conclusions conflicting with the answer.
    \item \textbf{Duplicate Removal:} If multiple chains are textually identical, keep only the first occurrence.
    \item \textbf{Order \& Priority:} Preserve the original order of appearance in \texttt{original\_reasoning}. If more than 10 chains remain, output the 10 most useful ones for deriving the answer (strongest/direct connection).
    \item \textbf{Text Integrity:} Do not modify any retained reasoning chain text. Each chain must remain exactly identical to its original text (from \texttt{'A->B->C->...'} to the end). Only reassign new indices starting from 1 in ascending order.
    \item \textbf{Empty Case:} If no chains satisfy the rules, output nothing (no text, no comment).
\end{enumerate}

\vspace{0.2cm}
\hrule
\vspace{0.2cm}

\textbf{\textsc{User Input}}

Input:
\begin{itemize}[leftmargin=*, nosep]
    \item \textbf{question:} \{question\}
    \item \textbf{answer:} \{answer\}
    \item \textbf{original\_reasoning} (each line formatted as \texttt{"<index>: A->B->C->..."}): \{original\_reasoning\}
\end{itemize}

Output:

Return only the filtered reasoning chains in this exact format and nothing else:
\begin{verbatim}
1: reasoning chain text (identical to original)
2: reasoning chain text (identical to original)
...
(Up to 10 lines total.)
\end{verbatim}

\end{tcolorbox}
\label{tab:phase2_prompt}
\end{table*}

\begin{table*}[t]
\centering
\begin{tcolorbox}[
  title={Phase 2: Reasoning Chain Refinement Template},
  colback=gray!5,
  colframe=black!75,
  fonttitle=\bfseries,
  width=\textwidth,
  boxrule=0.4pt
]
\small

\textbf{\textsc{System Prompt}}

You are an expert in the medical domain.

\textbf{Goal:} Generate reasoning chains where the \textbf{Start Node} is strongly correlated with a key entity in the Question, and the \textbf{End Node} is strongly correlated with the Answer entity.

\textbf{STRONG PRIORITY ON USING PROVIDED PATHS:}
\begin{itemize}[leftmargin=*, nosep]
    \item Maximize reuse of entities and links that appear in the provided Paths.
    \item Prefer chains composed entirely of nodes from the Paths.
    \item If you reuse nodes from the Paths, keep their strings EXACTLY as written (same casing, no edits).
    \item Select Start/End nodes from the provided Paths that have the highest semantic or clinical correlation to the Question/Answer context.
\end{itemize}

\textbf{STRICT OUTPUT RULES:}
\begin{enumerate}[label=\arabic*), leftmargin=*, nosep]
    \item \textbf{Format:} Each line: \texttt{'<index>: A->B->C->...'} (Index starts at 1, exactly one space after colon).
    \item \textbf{Delimiter:} Use \texttt{'->'} as the ONLY delimiter with no spaces around it.
    \item \textbf{Start Node:} Must have a STRONG CORRELATION (semantic or clinical) to a key entity in the Question (does NOT need to be an exact string match).
    \item \textbf{Final Node:} Must have a STRONG CORRELATION to the Answer entity (does NOT need to be the exact Answer string).
    \item \textbf{Validity:} Each link A$\to$B must be a medically valid causal/inferential relation.
    \item \textbf{Cleanliness:} No headers, no explanations, no extra text. Do not rewrite/rename nodes taken from the Paths.
    \item \textbf{Quantity:} Output up to 6 chains; output nothing if no valid chain can be formed.
    \item \textbf{Preference:} If multiple valid options exist, prefer chains that (a) maximize coverage of nodes from the provided Paths, and (b) require zero new nodes (or at most one new medically sound bridge).
    \item \textbf{Diversity:} Avoid producing only a single reasoning chain unless only one medically valid path exists; whenever possible, output two or more distinct valid reasoning chains.
\end{enumerate}

\vspace{0.2cm}
\hrule
\vspace{0.2cm}

\textbf{\textsc{User Input}}

\begin{itemize}[leftmargin=*, nosep]
    \item \textbf{Question:} \{question\}
    \item \textbf{Answer:} \{answer\}
    \item \textbf{Paths (filtered reasoning paths to reuse):} \{filter\_reasoning\_path\}
\end{itemize}

Output:
\begin{verbatim}
1: A->B->...
2: A->B->...
...
\end{verbatim}

\end{tcolorbox}
\label{tab:phase2_refinement_prompt}
\end{table*}

\begin{table*}[t]
\centering
\begin{tcolorbox}[
  title={Phase 2: Reasoning Chain Editing Template},
  colback=gray!5,
  colframe=black!75,
  fonttitle=\bfseries,
  width=\textwidth,
  boxrule=0.4pt
]
\small

\textbf{\textsc{System Prompt}}

You are a medical-domain reasoning chain editor.

\textbf{Edit policy (hard constraints):}
\begin{itemize}[leftmargin=*, nosep]
    \item \textbf{Identity-by-default:} If a chain is already complete and logically sound, output it UNCHANGED.
    \item \textbf{Only-if-incomplete:} Modify a chain ONLY when it is logically incomplete between Question-entity and Answer-entity/synonym.
    \item \textbf{Preserve-original-entities:} Do NOT alter, delete, paraphrase, or reorder ANY existing entities or links.
    \item \textbf{Insert-only-new-bridge:} When a fix is necessary, INSERT the FEWEST possible concise medical entities as bridges; do not modify existing tokens. Prefer $\le$ 2 new entities per chain.
    \item \textbf{Medical validity:} Each hop A$\to$B must be a clinically valid causal/inferential relation.
    \item \textbf{Uncertainty:} If uncertain whether a change is required, leave the chain UNCHANGED.
\end{itemize}

\vspace{0.2cm}
\hrule
\vspace{0.2cm}

\textbf{\textsc{User Input}}

\textbf{Requirements:}
\begin{enumerate}[label=\arabic*), leftmargin=*, nosep]
    \item Only add new entities to chains that are logically incomplete; keep ALL original entities exactly as given (no edits, no reordering, no deletions).
    \item Use \texttt{'->'} only; no spaces around it. Output one line per chain as \texttt{'<index>: A->B->C->...'}; indices start at 1, increment by 1, and have exactly one space after \texttt{':'}.
\end{enumerate}

\vspace{0.1cm}
\textbf{Input Data:}
\begin{itemize}[leftmargin=*, nosep]
    \item \textbf{Question:} \{question\}
    \item \textbf{Answer:} \{answer\}
    \item \textbf{new\_reasoning\_path:} \{new\_reasoning\_path\}
\end{itemize}

Output:
\begin{verbatim}
1: A->B->...
2: A->B->...
...
\end{verbatim}

\end{tcolorbox}
\label{tab:phase2_editing_prompt}
\end{table*}

\begin{table*}[t]
\centering
\begin{tcolorbox}[
  title={Phase 3: Atomic Reasoning Step Template},
  colback=gray!5,
  colframe=black!75,
  fonttitle=\bfseries,
  width=\textwidth,
  boxrule=0.4pt
]
\small

\textbf{\textsc{System Prompt}}

You are an expert in the medical domain. Your task is to perform \textbf{Chain-of-Thought (CoT) reasoning for a single step independently}, strictly based on prior dependency results and current step keywords.

\textbf{Core Task:}
Reason through \textbf{only the current step}. Rely solely on:
1) The CoT results from the directly dependent previous steps.
2) The entity keywords specific to the current step.

\textbf{Strict Constraints (Do NOT):}
\begin{itemize}[leftmargin=*, nosep]
    \item Do not explain the overall goal or speculate about future steps.
    \item Do not explicitly mention "dependencies" or reference previous steps by name (e.g., "based on step 1").
    \item Do not introduce extraneous definitions or general knowledge unrelated to the specific structural logic.
    \item Do not state functions or importance; stick to factual, anatomy-based structural logic.
\end{itemize}

\textbf{Mandatory Requirements:}
\begin{itemize}[leftmargin=*, nosep]
    \item \textbf{Conciseness:} Output only a single, short CoT paragraph.
    \item \textbf{Entity Detail:} All entities mentioned in the current step AND its dependencies \textbf{MUST} have their factual details explicitly included (e.g., structural, anatomical, or pathological attributes).
    \item \textbf{completeness:} The reasoning is considered incorrect if any entity detail is omitted or vaguely referenced.
\end{itemize}

\vspace{0.2cm}
\hrule
\vspace{0.2cm}

\textbf{\textsc{User Input}}

\textbf{Input Data:}
\begin{enumerate}[label=\arabic*., leftmargin=*, nosep]
    \item \textbf{Goal:} \{goal\}
    \item \textbf{Plan} (Steps \& Dependencies): \{plan\}
    \item \textbf{Executed Steps} (CoT results from dependencies only): \{executed\_step\}
    \item \textbf{Current Step} (Focus of this reasoning): \{current\_step\}
\end{enumerate}

\textbf{Output Instruction:}
\textbf{CoT Paragraph:} (Provide only one short paragraph of factual reasoning. Do not reference the answer or future steps.)

\end{tcolorbox}
\label{tab:phase3_execution_prompt}
\end{table*}

\begin{table*}[t]
\centering
\begin{tcolorbox}[
  title={Phase 3: Reasoning Refinement Template},
  colback=gray!5,
  colframe=black!75,
  fonttitle=\bfseries,
  width=\textwidth,
  boxrule=0.4pt
]
\small

\textbf{\textsc{System Prompt}}

You are an expert in extracting concise, non-redundant, factual reasoning from multi-step medical analyses.

\textbf{Input Context:}
You will be given:
1) A \textbf{goal}, describing the final medical question.
2) A series of \textbf{step-wise reasoning outputs}, where each step includes a label (\texttt{Entity A -> Entity B}) and a detailed reasoning paragraph.

\textbf{Task:}
Eliminate any redundant content that has already appeared in previous steps. Keep only factual details necessary for reasoning toward the goal.

\textbf{Strict Constraints:}
\begin{itemize}[leftmargin=*, nosep]
    \item \textbf{No Redundancy:} Remove information repeated from previous steps.
    \item \textbf{Facts Only:} Contain only medical facts; exclude definitions, purpose, significance, or usefulness.
    \item \textbf{Logic Preservation:} Must retain the logical relationship between Entity A and Entity B.
    \item \textbf{Format:} Return the original step label followed by your revised concise reasoning.
    \item \textbf{No Meta-Content:} Do not include explanations, summaries, or overall conclusions.
\end{itemize}

\vspace{0.2cm}
\hrule
\vspace{0.2cm}

\textbf{\textsc{User Input}}

\begin{itemize}[leftmargin=*, nosep]
    \item \textbf{Goal:} \{goal\}
    \item \textbf{Stepwise Reasoning:} \{multistep\_reasoning\}
\end{itemize}

\end{tcolorbox}
\label{tab:phase3_dedup_prompt}
\end{table*}

\begin{table*}[t]
\centering
\begin{tcolorbox}[
  title={Phase 3: Conclusion Synthesis Template},
  colback=gray!5,
  colframe=black!75,
  fonttitle=\bfseries,
  width=\textwidth,
  boxrule=0.4pt
]
\small

\textbf{\textsc{System Prompt}}

You are an expert in medical reasoning. You will be given three inputs:
1) Reasoning Paragraphs (a sequence of \texttt{'Transient Step N: ...'} logical steps),
2) Question (a medical multiple-choice question),
3) Options (possible answers).

\textbf{Your Task:}
\begin{itemize}[leftmargin=*, nosep]
    \item Use \textbf{ONLY} the Reasoning Paragraphs to determine the correct answer.
    \item First, output the final answer to the Question.
    \item Then, output one concise paragraph explaining why this is the correct answer, referencing the relevant steps.
    \item \textbf{Constraint:} Do not add external knowledge.
\end{itemize}

\textbf{Output Format:}
\begin{verbatim}
Explanation: <one paragraph justification>
Answer: <the correct option>
\end{verbatim}

\vspace{0.2cm}
\hrule
\vspace{0.2cm}

\textbf{\textsc{User Input}}

\begin{itemize}[leftmargin=*, nosep]
    \item \textbf{Reasoning Paragraphs:} \{total\_final\}
    \item \textbf{Question:} \{question\}
    \item \textbf{Options:} \{option\}
\end{itemize}

\end{tcolorbox}
\label{tab:phase4_conclusion_prompt}
\end{table*}

\begin{table*}[t]
\centering
\begin{tcolorbox}[
  title={Phase 4: Answer \& Logic Verification Template},
  colback=gray!5,
  colframe=black!75,
  fonttitle=\bfseries,
  width=\textwidth,
  boxrule=0.4pt
]
\small

\textbf{\textsc{System Prompt}}

You are an expert evaluator of medical reasoning consistency.

\textbf{Input Context:}
You will be given:
1) A \textbf{goal} (the correct answer).
2) A \textbf{question} and its \textbf{options}.
3) A \textbf{conclusion} (final answer + explanation).
4) A set of \textbf{reasoning steps} (the chain of thought).

\textbf{Your Tasks:}
\begin{enumerate}[label=\arabic*., leftmargin=*, nosep]
    \item \textbf{Answer Verification:} Verify whether the conclusion's final answer matches the correct answer specified in the goal.
    \item \textbf{Logic Verification:} Verify whether the explanation in the conclusion is logically consistent with the reasoning steps.
    \begin{itemize}[leftmargin=*, nosep]
        \item The explanation must be directly derivable from the given reasoning steps.
        \item It must \textbf{not} rely on external facts or background knowledge not present in the reasoning steps.
    \end{itemize}
\end{enumerate}

\textbf{Output Requirements:}
\begin{itemize}[leftmargin=*, nosep]
    \item Output \textbf{"Consistent"} if AND ONLY IF: (a) the answer matches the goal AND (b) the explanation is logically derived solely from the steps.
    \item Otherwise, output \textbf{"Inconsistent"}.
    \item Do not output anything else.
\end{itemize}

\vspace{0.2cm}
\hrule
\vspace{0.2cm}

\textbf{\textsc{User Input}}

\begin{itemize}[leftmargin=*, nosep]
    \item \textbf{Goal (correct answer):} \{goal\}
    \item \textbf{Question:} \{question\}
    \item \textbf{Options:} \{options\}
    \item \textbf{Conclusion (explanation + answer):} \{conclusion\}
    \item \textbf{Reasoning Steps:} \{reasoning\_steps\}
\end{itemize}

\end{tcolorbox}
\label{tab:phase4_verification_prompt}
\end{table*}

\section{Use of Large Language Models}
\label{app:llm_use}

We utilized large language models (LLMs) exclusively for linguistic refinement and as coding assistants for routine programming tasks. These tools played no role in the study's conceptualization, experimental design, data analysis, or the interpretation of result. The authors retain full responsibility for the validity and originality of the scientific content.

\end{document}